\documentclass{article}

\usepackage{microtype}
\usepackage{graphicx}
\usepackage{subfigure}
\usepackage{booktabs} 

\usepackage{hyperref}

\usepackage[accepted]{icmlw2019generalization}

\icmltitlerunning{Factors for the Generalisation of Identity Relations by Neural Networks}

\begin{document}

\twocolumn[
\icmltitle{Factors for the Generalisation of Identity Relations by Neural Networks}



\begin{icmlauthorlist}
\icmlauthor{Radha Kopparti}{to}and  
\icmlauthor{Tillman Weyde}{to}

\icmlcorrespondingauthor{Radha Kopparti}{radha.kopparti@city.ac.uk}

\end{icmlauthorlist}

\icmlaffiliation{to}{Research Centre for Machine Learning, Department of Computer Science, City University of London, UK }


\vskip 0.3in
]



\printAffiliationsAndNotice{}  

\begin{abstract}
Many researchers implicitly assume that neural networks learn 
arbitrary %
relations and generalise them to new unseen data. 
It has been shown recently, however, that the generalisation of  feed-forward networks fails for identity relations. %
%
The proposed solution for this problem is to create an inductive bias with  Differential Rectifier  (DR) units. 
In this work we explore various  factors in the neural network architecture and learning process whether they make a difference to the generalisation on equality detection of Neural Networks without and and with DR units in early and mid fusion architectures. 

We find in experiments with synthetic data effects of the number of hidden layers, the activation function and the data representation.   
The training set size in relation to the total possible set of vectors also makes a difference. 
However, the accuracy never exceeds 61\% without DR units at 50\% chance level. 
DR units improve generalisation in all tasks and lead to almost perfect test accuracy in the Mid Fusion setting. 
Thus, DR units seem to be a promising approach for creating  generalisation abilities that standard networks lack. 

\end{abstract}

\section{Introduction}
Humans are very effective 
in 
learning abstract patterns
that don't depend on the values of items but only on their relations, 
such as the identity 
of pattern components, 
and applying them to new stimuli, often after very brief exposure to a sequence of patterns \cite{GaryMarcus1999}. 
There, it was also shown that recurrent neural networks do not  generalise identity rules. 
This has recently been confirmed for the most common neural network types and the  `Relation Based Patterns' (RBP) solution, based on  `Differentiator-Rectifier' (DR) units has been introduced in \cite{nesy_2018}. 
The more fundamental problem of learning equality relations has identified and a solution with DR units has been proposed in \cite{r2l-2018}.



In this work, we study the generalisation behaviour of neural networks, especially feed-forward neural networks (FFNN) on the equality detection task. 
FFNNs are universal approximators \cite{leshno1993multilayer} and thus the modelling of equality between binary vectors is in their hypothesis space. 
The failure to generalise is therefore in the learning process and we investigate the effect of the learning parameters on the generalisation of FFNNs with and without DR units. 



The remainder of this paper is organised as follows: Section \ref{sec-related} presents a literature review, including a description of the RBP/DR structures and the task of abstract rule learning.
Section \ref{sec:exp-res} presents the experimental results and discussion, and  the conclusions of this paper follow in  Section \ref{sec:conclusions}.

\section{Related work}\label{sec-related}
The question of systematic learning of abstractions by neural networks has been posed already in \cite{fodor1988connectionism}.  
In a well-known study 
a recurrent neural network failed to distinguish abstract patterns, based on equality relations between sequence elements, although seven-month-old infants 
could %
distinguish them after a few minutes of exposure \citep{GaryMarcus1999}.
This was followed by an exchange on rule learning by neural networks and in human language acquisition with different approaches to training and designing neural networks
\citep{Elman1999, Altmann2, Shultz1,vilcu2001generalization,vilcu2005two,shultz2006neural,vilcu2005two,Shastri-1999-spatiotemporal,Dominey,Alhama}.

A more specific problem of learning equality relations was posed in \cite{GaryMarcus2001} by showing that neural network learning of equality on even numbers in binary representation does not generalise to odd numbers.
\cite{mitchell2018extrapolation} addressed this problem with different approaches as an example for extrapolation and inductive biases for machine learning in natural language processing. 
%
Inductive biases 
can be realised in a number of ways and have 
attracted %
increased interest recently. 
They have been applied, for instance, 
to spatial reasoning \cite{hamrick2018relational}, to arithmetic tasks  \cite{trask2018neural} and to learning from few examples
\cite{snell2017prototypical}, suggesting a number of potential benefits.

RBP and DR units have been introduced in \cite{nesy_2018} and \cite{r2l-2018} 
as a  way to create an inductive bias for learning identity relations. 
DR units compare two input values by calculating the absolute difference: $f(x,y) = |x-y|$. 
Pairs of vectors are compared with one DR unit for every vector dimension with weights from the inputs to the DR units fixed at $1$. 
There are two ways of adding DR units to a standard neural network.
In \textit{Early Fusion}, DR units are concatenated to input units, and in \textit{Mid Fusion} they are concatenated to the first hidden layer. 
The DR units have activation $0$ for equal input and positive values otherwise.  
Learning the suitable summation weights for the DRs is sufficient for creating a generalisable equality detector, so that 
they make this learning task easier. 
They have been shown to have a positive effect on real-life tasks in  \cite{nesy_2018} and no adverse effects were found on other tasks in \cite{r2l-2018}.

\section{Experiments and Results}\label{sec:exp-res}



The task here is a simple one: detecting if two binary vectors $v_1,v_2$ with $n$ dimensions are equal. 
$v_1$ and $v_2$ are concatenated as input to a neural network.
We extend the work in \cite{r2l-2018} by experimenting with various factors to determine their effect on generalisation. 

Our plain FFNN is a network with one hidden layer and ReLU activation. 
Throughout our experiments, the network is  trained for 20 epochs, which led to convergence in all cases and 100\%  average training accuracy.  
We run 10 simulations for each configuration and round accuracy averages. 
We test the significance of differences between models with a Wilcoxon Signed Rank Test with threshold $p=.05$, over the results of the simulations, which does not assume normal distribution and tests for different medians.

\subsection{Variations of the data size}
\paragraph{Vector dimensionality}
We generate pairs of random binary vectors with dimensionality $n$ between 2 and 100 as shown in table~\ref{tab:eval2}.
For $n<10$ we use all the possible binary vectors to generate equal pairs, i.e. $2^n$ equal pairs, and random vectors for the unequal pairs. 
For $n\leq{}10$ we use a random, class balanced selection of 10000 vectors. 
We randomly generate the same number of  unequal vector pairs. 
Then we use stratified sampling to split the data 75:25 into train and test set.
The results are shown in Table \ref{tab:eval2}.

We see that the standard FFNNs barely exceed chance level (50\%). 
The early fusion model improves results, but never reaches full generalisation. 
The Mid Fusion reaches close to perfect test performance.
We tested for significant differences between the lowest dimensionalities ($n=2,3$) vs the highest ($n=90,100$), with a sample size of 20 each. 
We found significant differences only for the DR Early and Mid Fusion ($p<.01$), where accuracy is better for lower dimensionalities, but not for the plain FFNN. 

\begin{figure}[tb]
\centerline{\includegraphics[width=9cm]
{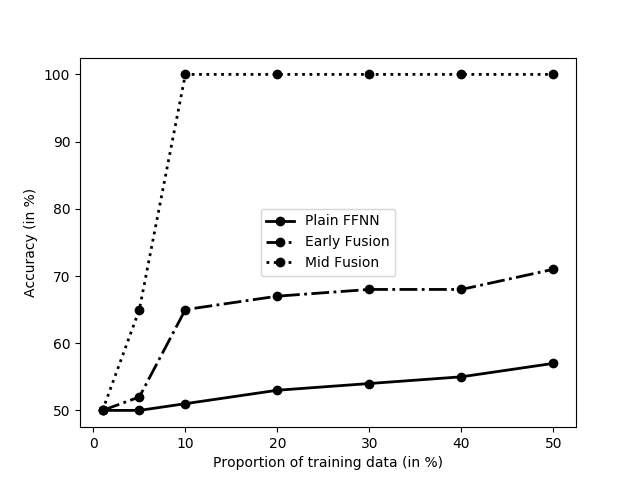}}
\caption{Accuracy of plain FFNN and DR variants for 10-dimensional binary vectors when varying the distributions of training data from 1\%  to 50\%, keeping the testing data fixed at 50\%.
}
\label{fig:var}
\end{figure}

\begin{table}[tb]
 \begin{center}
 \begin{tabular}{|l|r|r|r|}
  \hline
  Vector &  Plain  & DR Early  & DR Mid \\
    Dimension &  FFNN &  Fusion &  Fusion\\\hline

n=2 & 52\% & 66\%  & 100\% \\
n=3 & 55\% & 65\% & 100\% \\
n=5 & 51\% & 67\% & 100\% \\ 
n=10 & 51\% & 65\% & 100\% \\ 
n=20 & 49\% & 63\% & 100\% \\
n=30 & 50\% & 65\% & 100\% \\
n=40 & 50\% & 64\%  & 100\% \\
n=50 & 51\% & 65\% & 100\% \\
n=60 & 48\% & 64\% & 100\% \\
n=70 & 50\% & 62\% & 100\% \\
n=80 & 50\% & 63\% & 100\% \\
n=90 & 49\% & 62\% & 100\% \\
n=100 & 50\% & 64\% & 100\% \\
\hline
SD & 1.59 & 1.23 & 0.07 \\
\hline
 \end{tabular}
\end{center}
 \caption{Accuracy of the different network types on 
 pairs of vectors of different dimensions. 
 The joint train and test data covers all possible equal vector pairs for $n<10$, and a random, 
 class-balanced selection 
 of 10000 vector pairs where $n\ge 10$. 
 The standard deviation (SD) is given in percentage points.
 }
 \label{tab:eval2}
\end{table}
\paragraph{Training data size}
We study here how much the performance depends on the training data size.
For this, we vary only the training data size and keep the test set and all other parameters constant. 
We use training data sizes of 1\% to 50\% (in relation to the totally available data as defined above) and the accuracy achieved in various conditions is plotted in Figure~\ref{fig:var}. 
The Mid Fusion network reaches 100\% accuracy from 10\% data size on while the FFNN shows only small learning effects.

\begin{figure}[tb]
\centerline{\includegraphics[width=9cm]
{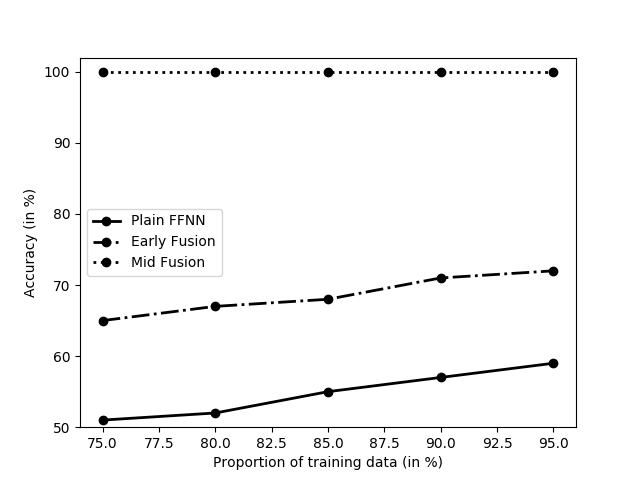}}
\caption{Accuracy of plain FFNN and DR variants for 10-dimensional vectors when varying the size of training data set from 75\% to 95\% of the total, with the remaining data used for testing.}
\label{fig:var4}
\end{figure}

In order to further study the effect of increased training data we also use higher proportions of the total data as training data, which means that we have to reduce the test data size.
We vary the train/test split from 75/25\% of training and testing data up to 95/5\% on pairs of 10-dimensional vectors. 
The results are given in Figure~\ref{fig:var4}, showing gradual improvements of the Early Fusion DR and plain FFNN models, but even with 95\% of all possible combinations in the training set the accuracy never exceeds 72\% and 68\%, respectively. 

\paragraph{Vector coverage}
A possible hypothesis for the results of the FFNN is that the coverage of the vectors in the training set plays a role. 
We do not share vectors in equal pairs between training and test set, as it would mean to train on the test data. 
Instead we create a training set that contains in its unequal pairs all vectors that appear in the test set. 
The results are shown in column a) of Table~\ref{tab:eval23}. 
We also created a training set where each vector appeared as above, 
but in two pairs, one in position $v_1$ and once in $v_2$. 
The results are shown in column b) of Table~\ref{tab:eval23}.
The results in both cases are similar to those without this additional coverage in Table~\ref{tab:eval2}.

\begin{table}[tbh!]
\begin{center}
\begin{tabular}{|l|r|r|r|}
\hline
  Type &  Plain  & DR Early  & DR Mid \\   
    & FFNN  & Fusion & Fusion \\ \hline
a) one & 50\% (1.56) &  75\% (1.17) & 100\% (0.03)\\
b) both & 52\% (1.56)  & 87\% (1.14) & 100\% (0.04)\\
\hline
\end{tabular}
\end{center}
\caption{Test set accuracy (standard deviation) of plain FFNNs and with DR units for test set vectors appearing in  training in a) one position, or b) both positions 
for $n=10$. 
Accuracies for a) and b) are not significantly different to the case of random vectors for plain FFNNs, but the improvement is significant for Early and Mid Fusion DR models.
}
\label{tab:eval23}
\end{table}

\subsection{Variations of the network architecture}

\paragraph{Network depth}
Given 
the general success of deep networks, we hypothesised that depth might help in learning a generalisable solution and thus tested networks with greater numbers of hidden layers. 
In the mid fusion architecture, the DR units are always concatenated to the first layer that is connected to the inputs. 
The results show some improvement, but not to a large extent, as we increase the number of hidden layers as given in Table~\ref{tab:eval8}. 
The difference between the shallow networks (1 and 2 layers) and the deep networks (4 and 5 layers) is statistically significant ($p<.01$).
This is an interesting result, as the additional depth is not necessary for representing a generalisable solution, but does help somewhat to find such a solution. 
It is not obvious, why deeper networks learn somewhat more generalisable solutions, given that Early Fusion models, where there is one more layer between the DR units and the outputs, are doing worse than the Mid Fusion models. 

\begin{table}[tbh!]
\begin{center}
\begin{tabular}{|l|r|r|r|}
\hline
 Number of  &  Plain & DR Early & DR Mid \\
 hidden layers & FFNN & Fusion & Fusion \\
\hline
h=1 & 55\% &  65\% & 100\% \\
h=2 & 57\% & 69\% & 100\%\\
h=3 & 58\% & 69\% & 100\%\\
h=4 & 60\%     &  72\%  & 100\%      \\
h=5 & 61\%     &  73\%    &   100\%     \\
\hline
SD & 1.57 & 1.59 & 0.05 \\
\hline
\end{tabular}
\end{center}
\caption{Test accuracy for different numbers of hidden layers with 10 neurons each, for vector dimension \textit{n=3}. 
Deeper networks are significantly better than shallower ones (see text for details).}
\label{tab:eval8}
\end{table}

\paragraph{Hidden layer width}
Based on the positive effect of a larger model with more parameters, we also evaluated the performance of the network using a single hidden layer and varying the number of neurons in that layer. 
We considered hidden layer size of 10 to 100, again with and without DR units, the results are tabulated in Table~\ref{tab:eval6}.
The observation here is that the larger models do not improve in performance when width is changed instead of depth. 
There is no significant difference between the two smallest networks ($h_n=10,20$) and the largest ($h_n=80,100$) for Plain FFNNs, while for the DR models the smaller networks perform significantly better. 

\begin{table}[tbh!]
 \begin{center}
 \begin{tabular}{|l|r|r|r|}
  \hline
  Hidden layer &  Plain  & DR Early  & DR Mid \\ 
  size & FFNN  & Fusion & Fusion \\ \hline
$h_{n}=10$ & 50\% & 65\% & 100\% \\ 
$h_{n}=20$ & 49\% & 63\% & 100\% \\ 
$h_{n}=30$ & 51\% & 61\% & 100\% \\
$h_{n}=40$ & 45\% & 65\% & 100\% \\ 
$h_{n}=50$ & 47\% & 65\% & 100\% \\ 
$h_{n}=80$ & 51\% & 62\% & 100\% \\ 
$h_{n}=100$ & 50\% & 64\% & 100\% \\
\hline
SD & 1.62 & 1.23 & 0.04 \\
\hline 
 \end{tabular}
\end{center}
 \caption{Accuracy of different network types for \textit{n=3}. The networks contain a single hidden layer of variable size $h_n$.}
 \label{tab:eval6}
\end{table}

\subsection{Other factors}

\paragraph{Activation function}
Another approach to change the learning behaviour is using different activation functions in the hidden layer. 
For \textit{n=3}, we evaluated the networks with ReLU, Sigmoid and Tanh activations. 
The results are given in Table~\ref{tab:eval4} and they show that the type of activation in the hidden layer has small positive effect in the overall accuracy, which is significant for all models. 

\begin{table}[tbh!]
\begin{center}
\begin{tabular}{|l|r|r|r|}
\hline
 Activation &  Plain  & DR Early  & DR Mid\\   
   function & FFNN  & Fusion & Fusion \\ \hline
ReLu & 55\% (1.38) &  65\% (1.23) & 100\% (0.05)\\
Sigmoid & 58\% (1.28) & 69\% (1.18) & 100\% (0.03)\\
Tanh & 58\% (1.29) & 69\% (1.17) & 100\% (0.03)\\
\hline
\end{tabular}
\end{center}
\caption{Test accuracy (standard deviation) of the network for different activation functions for vector dimension \textit{n=3}}
\label{tab:eval4}
\end{table}

\paragraph{Data representation}
We replace 0;1 with -1;1, like 
in \citep{courbariaux2016binarized}, but only applied to the inputs. 
The results of the accuracy for \textit{n=3} with and without DR units are given in Table~\ref{tab:eval5}.
We see an improvement, 
it is relatively small but significant across all models.

\begin{table}[tbh!]
\begin{center}
\begin{tabular}{|l|r|r|r|}
\hline
  Type &  Plain  & DR Early  & DR Mid\\   
    & FFNN  & Fusion & Fusion \\ \hline
a) 0/1 & 55\% (1.23) &  65\% (1.09) & 100\% (0.03) \\
b) -1/1 & 58\% (1.19) & 69\% (1.07) & 100\% (0.03)\\
\hline
\end{tabular}
\end{center}
\caption{Accuracy of the network for data representation. a) standard 0/1 representation and b) `sign' -1/1 representation for vector dimension \textit{n=3}.}
\label{tab:eval5}
\end{table}
We also tested a FFNN with 5 hidden layers, sigmoid activation function and -1/1 representation. Apparently the gains do not accumulate, as the test set accuracy was only 58\%.

\subsection{Other tasks}
Given the positive results for the DR mid fusion architecture for vector pairs, we want to test whether the DR units have an effect, possibly negative, on other learning tasks. 
We test a) comparison of the two vectors in the pair, b) checking if the digit sum is $\geq 3$, The results are shown in Table \ref{tab:eval3}.
In both a) and b) we see, that the performance is actually not hindered but helped by the DR units.  

We also test two tasks the DR architecture was not  designed for: c) digit reversal ($v_1 = flip(v_2)$) and d) parity checking (digit sum mod 2). 
The DR units organised per corresponding input neurons do not deliver a perfect solution here, but still lead to better results than a plain FFNN.  
The differences of Plain FFNN vs Early Fusion and Early vs Mid Fusion DR are statistically significant for all tasks. 

\begin{table}[tbh!]
\begin{center}
\begin{tabular}{|l|r|r|r|}
\hline
  Task &  Plain  & DR Early  & DR Mid\\   
& FFNN  & Fusion & Fusion \\ \hline
a)  & 75\% (1.08)  &  92\% (0.94) & 100\% (0.03)\\
b)  & 77\% (1.05) & 82\% (1.03) & 100\% (0.03)\\
c)  & 50\% (1.62) & 55\% (1.53) & 58\% (1.50)\\
d)  & 51\% (1.62) & 55\% (1.53) & 62\% (1.45)\\
\hline
\end{tabular}
\end{center}
\caption{Test set accuracy (standard deviation) of FFNN without and with DR units for a) numeric comparison ($v_1\geq v_2$), b) digit sum $\geq 3$, c) inversion of digit order and d) parity check.}
\label{tab:eval3}
\end{table}

\section{Conclusions} \label{sec:conclusions}
In this study we examined the lack of generalisation of identity rules by feed-forward neural networks 
and the effect of various factors in network architecture and learning method.
While data dimensionality, hidden layer size and vector coverage had no influence, the data representation, activation function and network depth do lead to some improvements.
Including DR units leads to substantial improvements. 
DR Mid Fusion reaches almost perfect generalisation, even from small amounts of training data, in all variants of identity detection. 

Identity is a fundamental task that many practitioners expect neural networks to solve. 
We therefore believe it is important to investigate the design of more techniques for creating and controlling inductive biases in neural network learning, as we find that even relatively a simple task like learning identity rules requires them for good generalisation. 

We see two sets of general questions that these results raise.
First: Why do FFNNs not learn to generalise vector identity? 
Which other relations do neural networks not learn? 
What does that mean for more complex tasks?
Second: What kinds of inductive biases should we design and how can we implement them for more complex tasks?

\bibliography{example_paper}
\bibliographystyle{icml2019}

\end{document}